\documentclass{article}

\PassOptionsToPackage{numbers, compress}{natbib}

\usepackage[preprint]{neurips_2020}

\usepackage[utf8]{inputenc} 
\usepackage[T1]{fontenc}    
\usepackage{hyperref}       
\usepackage{url}            
\usepackage{booktabs}       
\usepackage{amsfonts}       
\usepackage{nicefrac}       
\usepackage{microtype}      

\PassOptionsToPackage{hyphens}{url}\usepackage{hyperref}

\usepackage{times}  
\usepackage{helvet}  
\usepackage{courier}  
\usepackage{mathtools}

\usepackage{xspace}
\usepackage{latexsym}
\usepackage{amsfonts}
\usepackage{amsmath,bm}
\usepackage{amssymb}
\usepackage{color}
\usepackage{colortbl}
\usepackage{pbox}
\usepackage{array,multirow,graphicx}
\usepackage{paralist}
\usepackage{enumerate}
\usepackage[color,matrix,arrow,all]{xy}
\usepackage{comment}
\usepackage{booktabs}
\usepackage{balance}
\usepackage{stmaryrd}
\usepackage{pifont}
\usepackage{hhline}
\usepackage{listings}
\usepackage{array}
\usepackage{float}
\usepackage[flushleft]{threeparttable}
\usepackage{graphicx,wrapfig,lipsum}
 \usepackage[framemethod=TikZ]{mdframed}
\usetikzlibrary{arrows}
\usepackage{rotating}

\DeclareMathAlphabet{\pazocal}{OMS}{zplm}{m}{n}

\usepackage{array}
\newcolumntype{M}{>{\begin{varwidth}{2cm}}l<{\end{varwidth}}}
\newcolumntype{L}[1]{>{\raggedright\let\newline\\\arraybackslash\hspace{0pt}}m{#1}}
\newcolumntype{C}[1]{>{\centering\let\newline\\\arraybackslash\hspace{0pt}}m{#1}}
\newcolumntype{R}[1]{>{\raggedleft\let\newline\\\arraybackslash\hspace{0pt}}m{#1}}

\newcommand{\Ls}{\mathcal{L}}

\usepackage{array}
\usepackage{url}
\usepackage{longtable}
\usepackage{multirow}
\usepackage{xcolor}
\usepackage{algorithmic}
\usepackage[ruled, noline, longend, linesnumbered, boxed]{algorithm2e}
\usepackage{mdwlist}
\usepackage{subcaption}

\title{Filling out the missing gaps: Time Series Imputation with Semi-Supervised Learning}

\author{Karan Aggarwal\thanks{This author’s work was completed prior to his
affiliation with Amazon.} \\
Department of Computer Science\\
University of Minnesota\\
Minneapolis, MN, USA \\
\texttt{aggar081@umn.edu} \\
\And
Jaideep Srivastava \\
Department of Computer Science\\
University of Minnesota\\
Minneapolis, MN, USA \\
\texttt{srivasta@umn.edu} \\
}

\begin{document}

\maketitle

\begin{abstract}
 Missing data in time series is a challenging issue affecting time series analysis. Missing data occurs due to problems like data drops or sensor malfunctioning. Imputation methods are used to fill in these values, with quality of imputation having a significant impact on downstream tasks like classification. In this work, we propose a semi-supervised imputation method, ST-Impute, that uses both unlabeled data along with downstream task's labeled data. ST-Impute is based on sparse self-attention and trains on tasks that mimic the imputation process. Our results indicate that the proposed method outperforms the existing supervised and unsupervised time series imputation methods measured on the imputation quality as well as on the downstream tasks ingesting imputed time series. 
 \end{abstract}

\section{Introduction}
Time series are ubiquitous in most of the practical applications from meteorological forecasting, healthcare monitoring, to financial predictions.
While recent advances in deep learning have made a huge impact on the field, one of the most commonly seen issues with time-series data are missing values. Real world time series data is messy as it suffers from problems like sensor failures, data drops during transmission, malfunctioning sensors. 

Missing data can have a serious impact on downstream models for time series classification, or forecasting. Given the scarcity of labeled data, especially in domains like healthcare, it is not feasible to discard entire time series data. Hence, practitioners need to create models while handling missing time series data, appropriately. To counter this issue, various imputation methods have been proposed in the literature. 

Note, the main utility of imputation methods is to help increase the accuracy on downstream tasks like classification or regression. These tasks suffer if the imputation method used inserts a value which alters the distribution of the time-series, leading the downstream model to make errors. 

Time series imputation methods can be categorized into two: \emph{unsupervised} imputation methods and \emph{supervised} imputation methods. Unsupervised imputation methods learn statistical patterns in the observed time series to interpolate the missing values. Methods in classical machine learning and statistics literature are mostly based on nearest neighbors to missing values or spline fitting or using state space models~\cite{moritz2017imputets}. Recent methods~\cite{lipton2015learning,cao2018brits} using deep learning have been proposed to impute missing values, either use unsupervised learning~\cite{liu2019naomi, fortuin2019gp} or supervised learning from downstream tasks~\cite{cao2018brits}. Supervised learning based methods use the downstream tasks like time series classification, as the primary teaching signal while imputing the missing values. These supervised learning methods have shown state-of-the-art performance on imputation accuracy~\cite{cao2018brits}.

However, in practical settings, especially in domains like healthcare labeled data is limited. As labeling is expensive as well as sensitive in domains like healthcare where access to data is highly restricted due to regulatory concerns. Unlabeled data on the other hand is relatively easier to access due as access to collecting and storing data has become cheaper over the last decades.
With this observation, it is important to design imputation methods that can use both unlabeled and labeled data, \textit{i.e.}, semi-supervised methods. 

In this work, we propose a novel semi-supervised method, Sparse Transformer based Imputation (ST-Impute), to impute missing time series values. Transformers~\cite{vaswani2017attention} initially introduced in the area of natural language processing~\cite{devlin2018bert}, have significantly improved the performance of time series forecasting methods as well~\cite{li2019enhancing}. We modify the transformer architecture for time series imputation task by using diagonal self-attention masking and sparse activation functions. We train the model on the task of masked imputation modeling (MIM) by imputing artificially removed values to mimic the imputation task. Our modifications with the diagonal self-attention mask allows the model to train on time series reconstruction for non-missing values, alongside the supervised downstream task objective. ST-Impute improves imputation by 2\%-9\% over the most competitive baselines. To summarize, this work makes the following contributions:
\begin{itemize}
    \item We propose a novel semi-supervised learning algorithm for time-series imputation based on transformer architecture. 
    \item We make modifications to self-attention blocks by using a diagonal self-attention masking and sparse connections. This improves performance considerably over the vanilla transformer architecture.
    \item We propose a masked imputation modeling loss to mimic the imputation task.
    \item Our results on three public datasets show that our method beats baselines over the imputation task, as well as downstream tasks like time series classification/regression.
\end{itemize}

We organize the rest of the paper as follows. Section~\ref{sec3:related} places our work in the context of the existing literature. Section~\ref{sec3:background} discusses background concepts like self-attention. Section~\ref{sec3:methodology} describes ST-Impute architecture and training objectives in detail. Section~\ref{sec3:experiments} lays out our experimental settings. We present our results and analysis in Section~\ref{sec3:results}. Finally, we conclude in Section~\ref{sec3:conc}.

\section{Related Work}
\label{sec3:related}
In this section, we give a summary of the literature in the area that has tried to address the problem of imputing the missing values in time series. The literature is vast in this area, and we categorize methods based on classical and deep learning based methods. 

\subsection{Classical Methods}
Most of the methods in this category can be attributed to statistics based on the neighborhood of missing values~\cite{little2019statistical}. Simplest techniques deploy mean imputation or median imputation. Other commonly used local statistics deploy exponential moving average over time windows to impute the missing values. Further, some methods based on k-nearest neighbors have also been proposed~\cite{li2009dynammo, kreindler2006effects, batista2002study}. The idea here is to interpolate the valid observations and use them for imputation of the missing values. These involve assumptions using polynomial fits over the missing values based on previously observed time series values~\cite{fung2006methods}. 

Further, imputation methods involving using auto-regressive modeling like ARIMA or Seasonally adjusted ARIMA~\cite{velicer2003time, fung2006methods} have been used. These methods have been combined with state space modeling like Kalman filters~\cite{harvey1990forecasting}. ARIMA based methods have also been combined with anomaly detection by framing the problem as anomaly detection~\cite{zhang2017time} and imputing values while finding anomalies in the time series. Evolutionary algorithms have also been deployed for this problem~\cite{garcia2008missing}.

Expectation maximization based algorithms and graphical methods have also been proposed in the last decade~\cite{garcia2010pattern, ghahramani1994supervised, li2009dynammo}. They represent missing values with a latent variable and learn a transition matrix/sampling from the latent variable to fill out the missing values. 

Another common class of methods is matrix factorization based methods. These methods impute the missing values using Matrix Factorization based on the principle of collaborative filtering. Collaborative filtering is used in recommender systems to find the affinity of users with items based on the principle that if users that have similar taste will consume similar items. Using this principle, we can impute the ``affinity" of a user with an item even when we don't have any history of the user's interaction with that item. Hence, it can be viewed as imputing the missing values for a 2D matrix of users and items. Applying it to time-series would mean that time series that are similar in local patterns (time) would be used to fill up the missing values for a time series. Matrix factorization is used to reconstruct the correlations among the time-series to complete the missing values. Some notable attempts have been made in recent years~\cite{mei2017nonnegative, yu2016temporal, alquier2019matrix}. 

Out of the above methods, ImputeTS~\cite{moritz2017imputets} which implements the state space modeling with Kalman smoothing is the most effective method across a multitude of data-sets on the imputation task. Next, we discuss the deep learning methods for time series imputation.

\subsection{Deep Learning Methods}
Recently, there has been a lot of interest in using deep learning methods for time series imputation. These methods have shown state-of-the-art results over classical methods described in the previous section. 

 Some of the earliest deep learning methods~\cite{lipton2016directly, choi2016doctor} on time series classification/regression tasks in presence of missing data concatenated missing data timestamps as a feature instead of imputing the missing the values. This works on the idea that missingness has a pattern to itself which is helpful in the downstream supervised learning task. One of the earliest attempts was done using recurrent models by Che et al.~\cite{che2018recurrent}, termed as GRU-D. It utilizes GRUs as the base model and utilizes a masking mechanism for the missing values. They propose the concept of~\textbf{time lag} which is used in subsequent literature heavily. GRU-D models each time series value as a combination of existing value and a predicted value modeled by GRU. If the value is not missing, the predicted value is not back-propagated and opposite, if otherwise. This is done by using a mask function. The predicted values are initialized using the mean value of time series or last observations. Then, the model uses GRU to learn to compute (impute) these missing values. Further, they also deploy a decay function mechanism which tries to model the impact of other values on the missing time series values based on how far these values are from the missing value. Farther a value, less influence it exerts on the missing value. This decay is performed on the hidden state of GRU. Their method outperformed the existing classical methods on multiple datasets.
 
 GRU-I~\cite{luo2018multivariate} built on the GRU-D method by using Generative Adversarial Networks (GANs) as the base network to predict the missing values. Generator is used to learn the distribution of time series by generating missing time series values. The discriminator tries to classify time series randomly sampled from  samples generated by the generator or an actual time series to `real' or `fake'. Generator succeeds when it is able to fool the discriminator. At the time of inference, a generator is used to impute the missing values. Further variants of GANs using novel architectures like TS-
GAN~\cite{luo2018multivariate}, $E^2$GAN~\cite{luo2019e2gan}, NAOMI~\cite{liu2019naomi} have been proposed with promising results. 

BRITS~\cite{cao2018brits} proposed a supervised time series imputation method, exploiting a bi-directional recurrent neural network architecture. It relies completely on labeled data to train the imputation model. It gives a superior performance over the unlabeled methods mentioned above. 

CDSA~\cite{ma2019cdsa} uses cross-dimensional self-attention (CDSA) to impute missing values for geo-tagged data. While they use transformer architecture, their method is exclusively designed for geo-tagged spatiotemporal data, and not suitable for other domains. Additionally, our proposed ST-Impute can use both labeled and unlabeled data, and applicable to all time series domains.  
 



\section{Background}
\label{sec3:background}
In this section, we describe semi-supervised learning, self-attention mechanism, and sparse self attention as these concepts are used in our method afterwards. 

\subsection{Semi-supervised Learning}
Semi-Supervised learning is a technique to utilize unlabelled data while training a machine learning model on a supervised task. Semi-supervised learning's differentiation with unsupervised learning being that it is not trained agnostic to the supervised task at hand. Semi-supervised learning in general can be expressed as a combination of supervised learning loss on the labeled data along with a regularization from the unlabeled data. This regularization helps guide the supervised learning component to produce a better classifier which can be argued to be better as it has access to a higher density of points in the input subspace. Most of the prior imputation literature is based on using supervised learning methods~\cite{cao2018brits, lipton2015learning} or purely unsupervised~\cite{liu2019naomi}.

\subsection{Transformers and Self-Attention}
Transformer networks are a type of neural network that were first introduced in the field of natural language processing~\cite{vaswani2017attention}. They use \emph{self-attention mechanisms}, which allow different parts of the input to be considered in relation to their significance. This is done by having each part of the input interact with other parts of the input sequence within a local window to produce the final weights that are used to make a prediction. Unlike Recurrent Neural Networks (RNNs), Transformer networks do not take into account the order of the sequence in the input. It uses a triplet of $(K, Q, V)$: Key $K$, Query $Q$, and value $V$ to calculate self-attention. The query is created from the input by projecting it into a vector of dimension $d_k$, and the value is created in the same way. A scaled dot-product between Q and K is used to calculate attention scores with the softmax function. The dot-product is normalized by a square root of dimension size $d_k$, so that the dot product does not grow too large in magnitude,  to avoid small gradients. Each row $i$ in softmaxed dot product matrix represents the importance of observation at place $i$ from the other observations. This is further scaled by $V$. 

\begin{equation}
    \mathrm{SelfAttention(K, Q, V)} = \mathrm{Softmax} (\frac{KQ^T}{\sqrt{d_k}}) V
\end{equation}

$K$, $Q$, $V$ are scaled version of input by multiplying with learnable weight matrices, $\mathbf{W}_K$, $\mathbf{W}_Q$, and $\mathbf{W}_V$.



\subsection{Sparse Self-Attention}
Softmax has been the most commonly used function in machine learning methods. However, in self-attention architecture as discussed above, the softmax is done on a dense matrix that considers all the possible combinations in an observation sequence. As softmax takes into account each paired element of the matrix, it will assign non-zero weight to each element of the  matrix. This leads to unnecessary weights to meaningless connections. Hence, especially in case of long sequences, the most impactful pairwise connections get drowned by this distribution of non-zero weights by softmax activation. 

Sparsemax~\cite{martins2016softmax}, proposed as an alternative to softmax, to solve the above mentioned issue with softmax. This was further expanded by Laha et al.~\cite{laha2018controllable} to Sparsegen activation function. The key idea here is to turn these non-consequential pairwise connections to zero attention weights, while bumping the attention score of the important ones, \textit{i.e.,} a sparse attention distribution. 

In particular, Sparsegen-lin activation projects the attention scores $\mathbf{a} \in \mathbb{R}^n$ onto a probability simplex $\mathbf{p} \in \mathbb{R}^n$, along with a regularization coefficient $\lambda < $ 1:
\begin{equation}
    \mathrm{Sparsegen(\mathbf{a})} = \underset{\mathbf{p} \in \Delta^{n-1}}{\mathrm{argmin}}  ||\mathbf{p} - \mathbf{a} ||^2 - \lambda ||\mathbf{p}||^2 
\end{equation}

with, $\Delta^{n-1} = \{\mathbf{p} \in \mathbb{R}^n | \sum \mathbf{p} = 1, \mathbf{p} > 0\}$ enforcing constraints of probabilities summing to one and greater than non-zero. Note, the L2 norm with negative $\lambda$ regularization acts to actually assign larger probability values in $\mathbf{p}$, as the objective is to minimize the cost function above. Next, a thresholding step is done, where the probabilities lower than a dynamic threshold $\tau$ are truncated to zero, while redistributing the remaining probabilities. For more details refer to the paper~\cite{laha2018controllable}. 

We can use this Sparse Self-Attention by replacing the softmax with Sparsegen function:
\begin{equation}
    \mathrm{SparseSelfAttention(K, Q, V)} = \mathrm{Sparsegen} (\frac{KQ^T}{\sqrt{d_k}}) V
\end{equation}

Note, sparse (self-)attention and its variants has been used in NLP works with impressive results~\cite{cui2019fine} as well as time series forecasting~\cite{li2019enhancing}.

\section{ST-Impute}
\label{sec3:methodology}

In this section, we introduce our proposed model, Sparse Transformer Imputation (ST-Impute). We go over the architecture and the proposed training objectives we use to train the ST-Impute model.

\subsection{Model Architecture}
We only use the encoder network from the original transformer network as we are not doing a Seq2Seq generation unlike the transformer model. This is similar to the famous pre-trained model in NLP called BERT~\cite{devlin2018bert}.

The original transformer was designed to perform Seq2Seq tasks like machine translation. However, since the goal of this work is imputation, we use a time-series reconstruction based approach, similar to BERT~\cite{devlin2018bert}. In ST-Impute, we do three modifications: 1) Missing Value Mask; 2) Diagonal Self-Attention Masking; and  3) Sparse self-attention.

\paragraph{Missing Value Mask}: We concatenate the input time series ($\mathbf{T} \in \mathbb{R}^{n}$) with a mask, $\mathbf{M} \in \mathbb{R}^{n}$ to indicate which value is a missing value. Note, mask dimensions are the same as input time series and would extend in case of multivariate time-series as well. This is passed to the linear layer of the model with ReLU activation, which projects it in model dimensions before positional encoding step:
\begin{equation}
    \mathrm{Linear} = \mathrm{ReLU} ([\mathbf{T};\mathbf{M}] W_e + b_e)
\end{equation}

where, $W_e$ and $b_e$ are linear projection matrix weights and bias. The results goes to positional encoding, after which it is fed to the self-attention blocks, modified as described next.

\paragraph{Diagonal Self-Attention Masking}: Since we take a reconstruction based approach in this work, we found that using a vanilla transformer makes it trivial for the encoder to reconstruct the observed time series values as it has access to these values through self-attention, with the attention scores lighting up on the diagonal. Another thing to consider would be that we want the model to learn to construct a missing value from neighbors. Hence, if we make the model predict the observed values through the neighbors, that would help it learn the patterns in time-series data which would be beneficial on the imputation task. We take a leaf from the transformer model~\cite{vaswani2017attention} where the attention scores for future words are masked in the decoder. UniLM~\cite{dong2019unified} and CDSA\cite{ma2019cdsa}, show that masking out diagonal elements or certain off-diagonal elements improves transformer's accuracy on various NLP tasks and time series imputation tasks.
In order to avoid the trivial re-construction case, we use following diagonal mask ($\mathbf{DM}$):
\begin{equation}
\mathbf{DM}(i,j) =
\begin{cases}
    \begin{array}{lr}
        0, & \text{if } i \neq j\\
        -\infty, & \text{if } i= j
    \end{array}
\end{cases}
\end{equation}

The Self-attention looks like following with addition of Diagonal Mask:
\begin{equation}
    \mathrm{SelfAttention(K, Q, V)} = \mathrm{Softmax} (\frac{KQ^T}{\sqrt{d_k}} + \mathbf{DM}) V
\end{equation}

\begin{figure}
\includegraphics[scale=0.3]{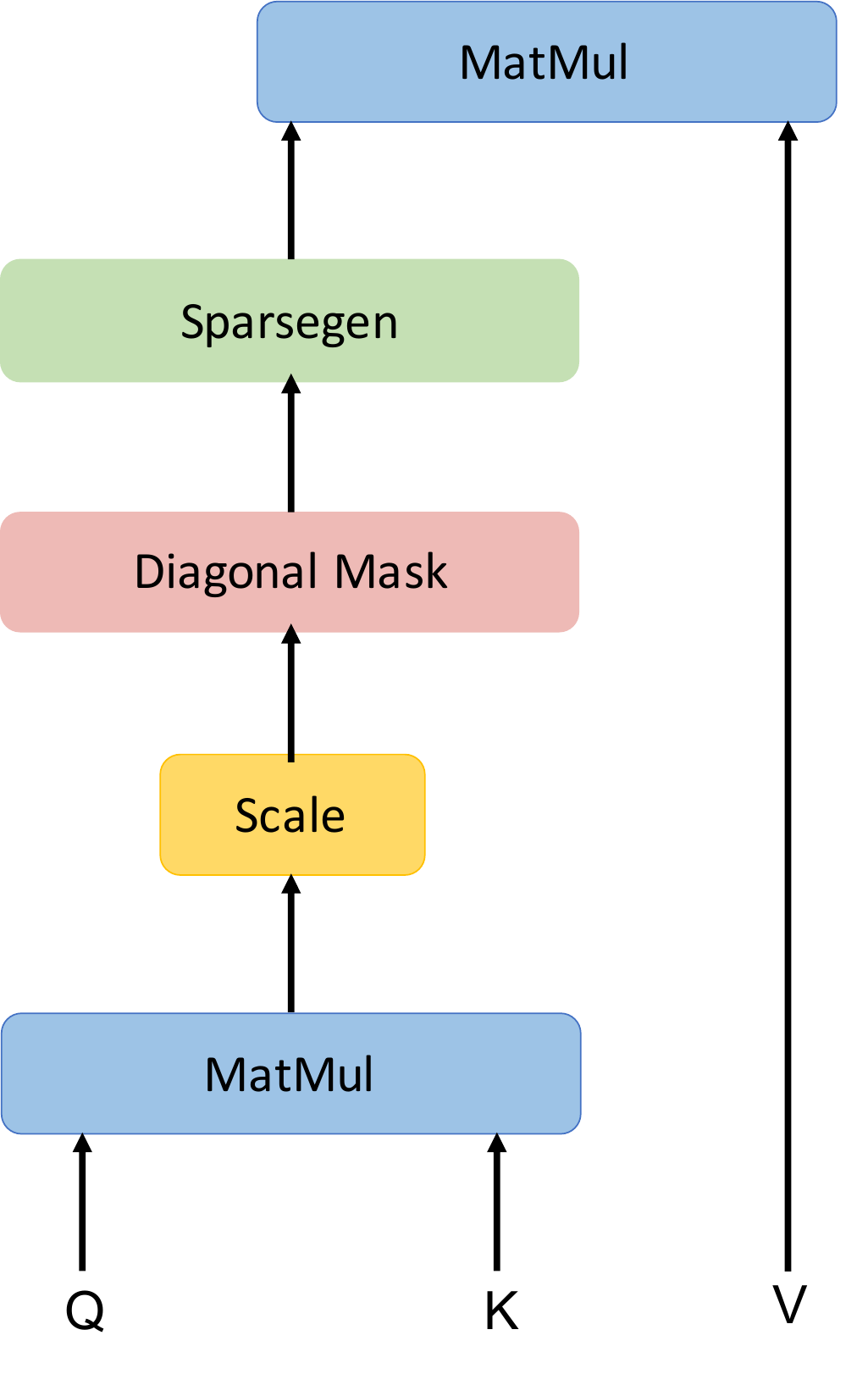}
\centering
\caption{Sparse self-attention block with diagonal self-attention masking. }
\centering
\label{fig:block}
\end{figure}

\paragraph{Sparse self-attention}: As described in the previous section, sparse self-attention has shown a noticeable performance increase over the traditional counterpart in NLP~\cite{cui2019fine}. We use sparse self-attention with diagonal self-attention masking which can be given by:

\begin{equation}
    \mathrm{SparseSelfAttention(K, Q, V)} = \mathrm{Sparsegen} (\frac{KQ^T}{\sqrt{d_k}} + \mathbf{DM}) V
\end{equation}

Figure~\ref{fig:block} shows the self-attention block used in ST-Impute.

\subsection{Training Objectives}
We describe three training objectives we use to train ST-Impute: 1) Masked Imputation Modeling; 2) Reconstruction Loss; and 3) Downstream task loss.

\paragraph{Masked Imputation Modeling}
Masked Imputation Modeling (MIM) is motivated by BERT~\cite{devlin2018bert}'s Masked Language Modeling task,  where the model is trained to predict randomly masked words. We mask the time series observations at random. The idea is for the model to predict these artificial missing values based on neighboring observed time series values. This forces the model to learn patterns which it can use to predict the missing values on real data, with feedback during the training phase. We use  Loss function for MIM task can be given as a Mean Absolute Error (MAE) on imputed values:
\begin{equation}
   \Ls_{\mathrm{MIM}} = {\sum_{i=0}^{n}} \frac{1}{\sum_{i=0}^{n}  M_i}|| (\mathbf{T} - \hat{\mathbf{T}}) \odot \mathbf{M}||
\end{equation}

where, $\mathbf{T}$ is original time series and $\hat{\mathbf{T}}$ is the estimated time-series, while $\mathbf{M}$ is imputation mask indicating the values that were randomly masked in the input.

\paragraph{Non-missing Reconstruction Loss}
Non-missing Reconstruction Loss (NRL) is added to the training for the model to accurately reconstruct the non-masked observed time series values. The idea here is for the model to learn the patterns in the time series data, which would help it converge faster as well as generalize better, instead of just learning from imputing the artificially removed values. This ensures the model does not stray away from the time-series patterns by overfitting on the MIM task. It also forces reconstruction of observed values with missing values in time-series  which mimics the imputation task. Hence, it is more sample efficient versus MIM alone, as the model learns time-series patterns to perform imputation in presence of missing values.  

\begin{equation}
   \Ls_{\mathrm{NRL}} = {\sum_{i=0}^{n}} \frac{1}{\sum_{i=0}^{n}  O_i}|| (\mathbf{T} - \hat{\mathbf{T}}) \odot \mathbf{O}||
\end{equation}

where, $\mathbf{O}$ is a mask indicating the values that were observed and were not removed during the MIM task. Note, $\mathbf{O}$ also excludes the time series values which were missing in the original data, and not artificially removed during the MIM task.

\paragraph{Semi-supervised Downstream task loss}
The two loss objectives for MIM and NRL, can work on both labeled as well as unlabeled data. We add downstream task loss for the cases where the downstream task (classification or regression) labels are available following previous works~\cite{cao2018brits}. In case of classification, it can be a simple cross-entropy loss while it can be a mean absolute error for a regression task. We add a single fully connected layer on top of ST-Impute output which then predicts the downstream task. This ensures that the downstream task guides the imputation learning process whenever labels are available in a \textbf{\emph{true semi-supervised learning fashion}}. This will be denoted by $\Ls_{\mathrm{c}}$.

\paragraph{Combined Training Loss}: Combined training loss function is given by:
\begin{equation}
   \Ls = \Ls_{\mathrm{MIM}} + \Ls_{\mathrm{NRL}} + \Ls_{\mathrm{c}}
\end{equation}


\section{Experimental Settings}
\label{sec3:experiments}
In this section, we present our experimental setup: datasets used, baselines used, and metrics used to measure the effectiveness of imputation and downstream tasks.  

\subsection{Datasets}
\label{c3:datasets}
We use the following datasets in this work, which have been used in prior works~\cite{che2018recurrent}:
\begin{itemize}
\item Physio-Net\footnote{\url{https://physionet.org/challenge/2012/}}: This multivariate clinical time series dataset contains 4000 examples  with about 41 physiological signals like heart rate, measured from intensive care unit (ICU) stays. Each patient has the data for the first 48 hours after ICU admission of the patient. Note, this dataset has one of the highest missing values --- about 80.67\% of values are missing. 10\% of the data is used as a test set following~\cite{cao2018brits}. The downstream task here is to predict in-hospital mortality.
\item Activity\footnote{\url{https://archive.ics.uci.edu/ml/datasets/Localization+Data+for+Person+Activity}}: This dataset contains multivariate time series
of motion state recording of 5 people. In total, 11 kinds of activities  were tracked like walking, lying, falling, sitting, or standing up. Four accelerometer sensors were attached to each person on the left/right ankle, chest, and belt to track the 3D coordinates. In total, 4,100 time series samples were used in the experiments over 40 consecutive time steps. 
\item KDD Cup\footnote{\url{https://www.microsoft.com/en-us/research/wp-content/uploads/2016/06/STMVL-Release.zip}}: is a public dataset containing PM2.5 particulate measurements from 36 monitoring stations in Beijing. These observations were collected hourly between 2014/05/01 to 2015/04/30. It has  365 time series samples with a label every 24 consecutive hours. It has a missingness rate of 13.30\%. We use the same train/test split as used in prior works~\cite{cao2018brits}: 3rd, 6th, 9th and 12th months are reserved as test data with the rest of nine months as training data. A consecutive observation of 36 samples is used as one time-series sample.
\end{itemize}

\subsection{Baselines Used}
We use the following baselines:
\begin{itemize}
\item \textbf{Mean Imputation}: Imputes the missing values with the mean value of the       time series.
\item \textbf{Last Imputation}: Imputes the missing value with the last seen value of the time-series.
\item \textbf{ImputeTS}~\cite{moritz2017imputets}: Kalman filtering based classical deep learning library for time series imputation. 
\item \textbf{BRITS}~\cite{cao2018brits}\footnote{\url{https://github.com/NIPS-BRITS/BRITS}}: Bi-directional LSTM based method that uses downstream class labels, and predicting the next value to guide the missing value prediction. The latter task is only computed on the present values of the time series. 
\item \textbf{GRU-D}~\cite{che2018recurrent}: This method is based on Gated Recurrent Units (GRUs), and use a flag/mask which indicates a missing value to train and impute time series.  
\item \textbf{GP-VAE}~\cite{fortuin2019gp}: This method uses the Gaussian Processes to guide the Variational Autoencoders (VAEs) to impute missing values. 
\item \textbf{NAOMI}~\cite{liu2019naomi}\footnote{\url{https://github.com/felixykliu/NAOMI}}:  Non-Autoregressive Multiresolution Imputation (NAOMI) uses a  divide-and-conquer strategy by doing imputation across multiple time resolutions, along with a generative adversarial training to impute missing values. This is an unsupervised learning based imputation method.
\end{itemize}

Note, BRITS and GRU-D can only be trained in the presence of labeled data as they need downstream task's labels for imputing. Hence, we only use the labeled data for them. For the other methods, we can use unlabeled data as well. 

\subsection{Hyper-parameters}
We use the same train-test split as provided in the datasets. For hyper-parameter tuning we randomly sample 80\% of the train set for training, and rest 20\% for validation. For transformer architecture, we try number of layers $\in \{1, 2, 3, 4, 5, 6, 7\}$, and selected 4 layers and 4 as the number of heads. For layer size, we searched $\in \{32, 64, 128, 256, 512\}$ and selected 128 with a dropout of 0.15. We used a learning rate of $0.0002$ with Adam optimizer.

\subsection{Imputation Quality Metrics}
Quality of imputation is measured by the following metrics:
\begin{itemize}
\item Root Mean Square Error (RMSE): Root mean square measures the average squared difference between the predicted and ground truth value. 
\end{itemize}

\subsection{Downstream Tasks}
Following two downstream tasks in these datasets that is used to measure the effectiveness of an imputation method:
\begin{itemize}
\item \textbf{Time Series Classification}: Goal is to predict the class of the time series, with most of the tasks being binary classification tasks. 
\item \textbf{Time Series Regression}: Goal is to predict a continuous target variable such as Air Quality Index. 
\end{itemize}

The hypothesis here being that if a method imputes a time series as close to the real time series, we would see better performance on these tasks. As the time series now contains more ``real" data which can be helpful for the downstream tasks. This hypothesis does assume that the missing parts of the time series do carry the information that is important for the downstream tasks.  

\section{Results and Analysis}
\label{sec3:results}
In this section, we present our results on our three public datasets described in Section~\ref{c3:datasets}. We present the results on imputation metrics with different missingness rates. Next, we show comparison on downstream tasks of regression and classification by using the data that is imputed with these methods. We also explore the effect of the amount of labeled data on imputation. We further explore the effect of missingness patterns in the ablation studies on our model.

\begin{table}[b]
    \centering
    \caption{RMSE of baseline imputation methods vs ours (ST-Impute) in terms of RMSE/MAE with different missing rates. Note, these results are with Missing at random.}
  \label{table:imputation_results}
  \centering
  \resizebox{\columnwidth}{!}{%
  \begin{tabular}
{|c|c|ccccccc|cc|}
\midrule
     \multicolumn{1}{|c|}{Dataset} & Missing Rate & Mean & Last & ImputeTS & GRU-D  & GP-VAE & NAOMI  & BRITS & Transformer & ST-Impute\\
     \midrule
     & 10\% & 0.812  & 0.792   & 0.702  & 0.722   & 0.670& 0.642 & 0.620 & 0.552 & \textcolor{blue}{\textbf{0.545}} \\
     & 30\% & 0.873 & 0.862 & 0.739 & 0.792   & 0.726  &  0.724 & 0.686  & 0.606 & \textcolor{blue}{\textbf{0.592}}\\
    Activity & 50\% & 0.933& 0.936 &  0.826 & 0.879 & 0.796 & 0.794 & 0.786  & 0.689 & \textcolor{blue}{\textbf{0.681}} \\
     & 70\% &  0.943 & 0.956 &  0.897 &  0.923  & 0.846 & 0.854 & 0.836 & 0.763 & \textcolor{blue}{\textbf{0.749}} \\
     & 90\% & 0.963 & 0.968  & 0.942 & 0.951 & 0.882  & 0.897 & 0.867  & 0.808 & \textcolor{blue}{\textbf{0.796}}\\
     \midrule
     & 10\% &  0.799 &0.802  & 0.707 & 0.710& 0.677 &0.632 &0.611 & 0.554 & \textcolor{blue}{\textbf{0.547}} \\
     & 30\% & 0.863 &0.855  & 0.759 & 0.782 &0.707 &0.703 &0.672  & 0.609 & \textcolor{blue}{\textbf{0.603}}\\
    PhysioNet & 50\% &  0.916 & 0.917 & 0.766 & 0.808 & 0.787 & 0.783 & 0.779 & 0.714 & \textcolor{blue}{\textbf{0.702}}\\
     & 70\% & 0.936 &0.947 & 0.827 &  0.848 &0.837& 0.835 & 0.809   & 0.729 & \textcolor{blue}{\textbf{0.720}}\\
     & 90\% &  0.952& 0.959  & 0.855 & 0.863 & 0.879 & 0.865& 0.850  & 0.772 & \textcolor{blue}{\textbf{0.764}} \\
     \midrule
    & 10\% & 0.763 & 0.789  & 0.609 & 0.701 & 0.522 &  0.522 & 0.531  & 0.432 & \textcolor{blue}{\textbf{0.430}} \\
     & 30\% & 0.806 & 0.804  & 0.651 & 0.761 & 0.562 & 0.558 & 0.561  & 0.457 & \textcolor{blue}{\textbf{0.453}} \\
    KDD & 50\% & 0.866 & 0.888 & 0.747 &  0.801 & 0.602  & 0.602 & 0.581  & 0.484 & \textcolor{blue}{\textbf{0.479}} \\
     & 70\% & 0.898 & 0.921  & 0.793 & 0.821 & 0.709 &  0.701&  0.641  & 0.596 & \textcolor{blue}{\textbf{0.589}} \\
     & 90\% & 0.919 & 0.949  & 0.833 & 0.842 & 0.771 &  0.762 & 0.720  & 0.654 & \textcolor{blue}{\textbf{0.653}}\\
    \bottomrule
\end{tabular}
}
\end{table}

\subsection{Imputation Performance versus Effect of missingness rate}
We study the effect of missingness rate on imputation performance of our method and baselines. For training, 50\% of the values in the training set are omitted at Missing Completely at Random, in line with prior works~\cite{cao2018brits}. For testing, we hold out 10\%-90\% of all the values present in the test set of the three datasets. Next, the trained imputation model is ran on the test set to impute the missing values. Imputation accuracy is calculated using RMSE on imputed values and real values that were held out. 

Imputation RMSE is reported in Table~\ref{table:imputation_results}. We can observe that our method outperforms all the baselines, including a purely Transformer based baseline. For all the methods, performance deteriorates as the missingness rate goes up from 10\% to 90\%, which is expected as the information used by all these methods (including ST-Impute) depends on the non-missing time series observations. As those drop, it becomes harder for the methods to recover the missing values. 

We can clearly see the progression in improved imputation from classical methods like ImputeTS to deep learning methods like BRITS, and ST-Impute. Further, ST-Impute has a consistently lower RMSE of 6\% - 10\% than BRITS. ST-Impute has considerable improvement over all three datasets, Activity, PhysioNet datasets, and KDD dataset. 

ST-Impute consistently outperforms a pure Transformer model as well - more so on Activity and PhysioNet datasets, while performing similarly for KDD dataset. This demonstrates that using sparsity helped significantly to improve the performance over a purely vanilla transformer.

\begin{table}
{
\caption{AUC-ROC and PR-AUC scores for PhysioNet data imputed with various imputation methods.}
\label{table:imputeClassResults}
\begin{center}
\begin{tabular}{|l|cc|} 
\toprule
\textbf{Method} & AUC-ROC & PR-AUC \\
\midrule
Mean & 83.3\% $\pm$ 0.3\% & 46.0\% $\pm$ 0.7\%  \\
Last & 82.7\% $\pm$ 0.3\% & 46.8\% $\pm$ 0.4\%\\
\midrule
GRU-D & 82.9\% $\pm$ 0.3\% & 45.6\% $\pm$ 0.7\%\\
GP-VAE & 83.4\% $\pm$ 0.2\% & 48.2\% $\pm$ 0.6\% \\
NAOMI & 83.5\% $\pm$ 0.4\% & 49.0\% $\pm$ 0.7\%\\
BRITS & 83.5\% $\pm$ 0.2\% & 49.2\% $\pm$ 0.3\%\\
\midrule
Transformer & 84.3\% $\pm$ 0.4\% & 49.8\% $\pm$ 0.8\%\\
ST-Impute & \textcolor{blue}{\textbf{84.7\%}} $\pm$ 0.3\% & \textcolor{blue}{\textbf{50.6\%}} $\pm$ 0.5\% \\
\bottomrule
\end{tabular}
\end{center}
}
\end{table}

\subsection{Downstream task: Classification}
We further evaluate our methods on classification task in the PhysioNet dataset. The task is to predict mortality of a patient, \textit{i.e.}, if a patient would die in the next 2 weeks given the physiological signals at the current time. Note, this dataset has 80\% missing values in the existing time-series which makes the predictions non-trivial on this dataset. In line with previous works~\cite{cao2018brits}, we use our method to impute the existing missing 80\% values in the dataset. Note, we do not add any new missing values, we just impute the missing values in the real dataset. It uses the same test set as reported earlier.

Next, the imputed time series are used to train a downstream classifier to predict patient mortality. The key idea here is that better an imputation method is, closer it is to real time series values. The core assumption here is that the closer the imputed time series is to real values, the better its performance is on the classification task. While the assumption may not hold true, it can provide a directional signal on quality of imputation as it would signify that the imputation method was able to capture patterns in the time series to be able to impute it correctly, helping the downstream classifier in the signals that can discriminate between time series classes.

For the classifier, we use a simple RNN classifier model as done by previous works~\cite{cao2018brits}. RNN model is trained and tested with imputed datasets. In line with previous works, we use 30 training epochs, a learning rate of 0.005, dropout of 0.5, and hidden state dimensions of 64 for RNN classifier. The RNN classifier architecture is identical for every imputed dataset so that we can compare the imputation models fairly. 

Results are presented in Table~\ref{table:imputeClassResults}. We observe that the classifier trained on data imputed by ST-Impute beats all the existing methods on the classification task on both AUC-ROC and PR-AUC metrics. Since, the dataset is imbalanced - 15\% of labels has a mortality, PR-AUC is a better metric. We observe an increase of 1.3\% on AUC-ROC and an increase of 2.7\% on PR-AUC metric with ST-Impute versus the next best method, BRITS. Further, we observe that ST-Impute improves considerably over the pure transformer based baseline. We can attribute this to the fact that this dataset has about 80\% existing missing values, making the imputation task quite tricky. With a Transformer like model, there is a chance of over fitting on the sparse observed values.  Using sparsity with Transformer, helps reduce this over fitting and increase the downstream task performance. 

\subsection{Downstream Task: Regression}
Identical to the classification task setting, we evaluate the imputation quality using a downstream regression task on KDD dataset. Instead of using a classification layer, we use a regression layer with the same architecture. We use identical training settings as the classification task above, for fairness sake as used in previous works~\cite{cao2018brits}. 

\begin{figure}
\includegraphics[width=\linewidth]{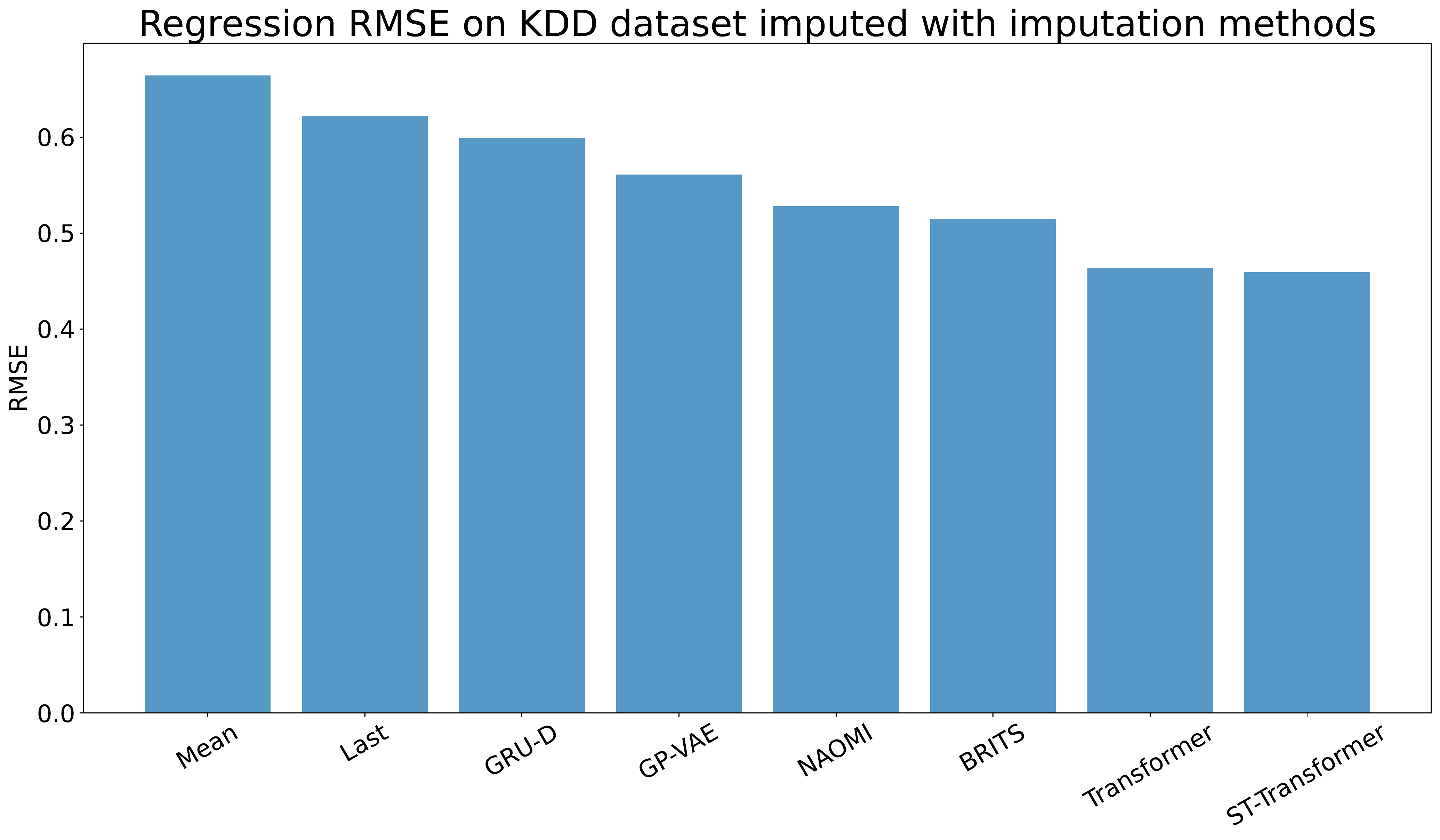}
\centering
\caption{ RMSE for Regression on KDD dataset (50\% missingness) imputed with various imputation methods.}
\centering
\label{fig:KDDregression}
\end{figure}

Results are shown in Figure~\ref{fig:KDDregression}. We observe that the regression model trained on data imputed by ST-Impute has the lowest RMSE on the regression task. ST-Impute has a 5.1\% lower RMSE than BRITS, and a 0.9\% lower RMSE than the Transformer model, consistent with previous showing for the imputation task and classification tasks. While performance on downstream tasks is an indirect way of measuring algorithm's performance, this shows that ST-Impute's imputations are closer to real world unobserved values. We can claim that increased imputation accuracy helps boost performance of downstream regression task.

\begin{table}[b]
\centering
\caption{Impact of Loss Functions and Masking on imputation for Activity Dataset.}
\label{table:ablation}
\scalebox{1}{\begin{tabular}{|l|c|} 
\toprule
\textbf{Method} & Imputation RMSE  \\
\midrule
ST-Impute (-NRL) & 0.684\\
ST-Impute (-Classification Loss) & 0.767\\
ST-Impute (-Diagonal Self-Attention Mask) & 0.698\\
ST-Impute (-Sparsegen) [Transformer] & 0.689\\
ST-Impute & \textbf{0.681} \\
\bottomrule
\end{tabular}
}
\end{table}

\subsection{Ablation study: Impact of Loss Functions and Masking}
We perform an ablation study by removing certain aspects of ST-Impute to show their impact on imputation performance with Activity Datasets (50\% missing rate). Results are shown in Tabel~\ref{table:ablation}. As we can observe, the most significant impact is from diagonal self-attention mask and removing the label data information through classification loss, while NRL a marginal difference.

\begin{figure}
\includegraphics[width=\linewidth]{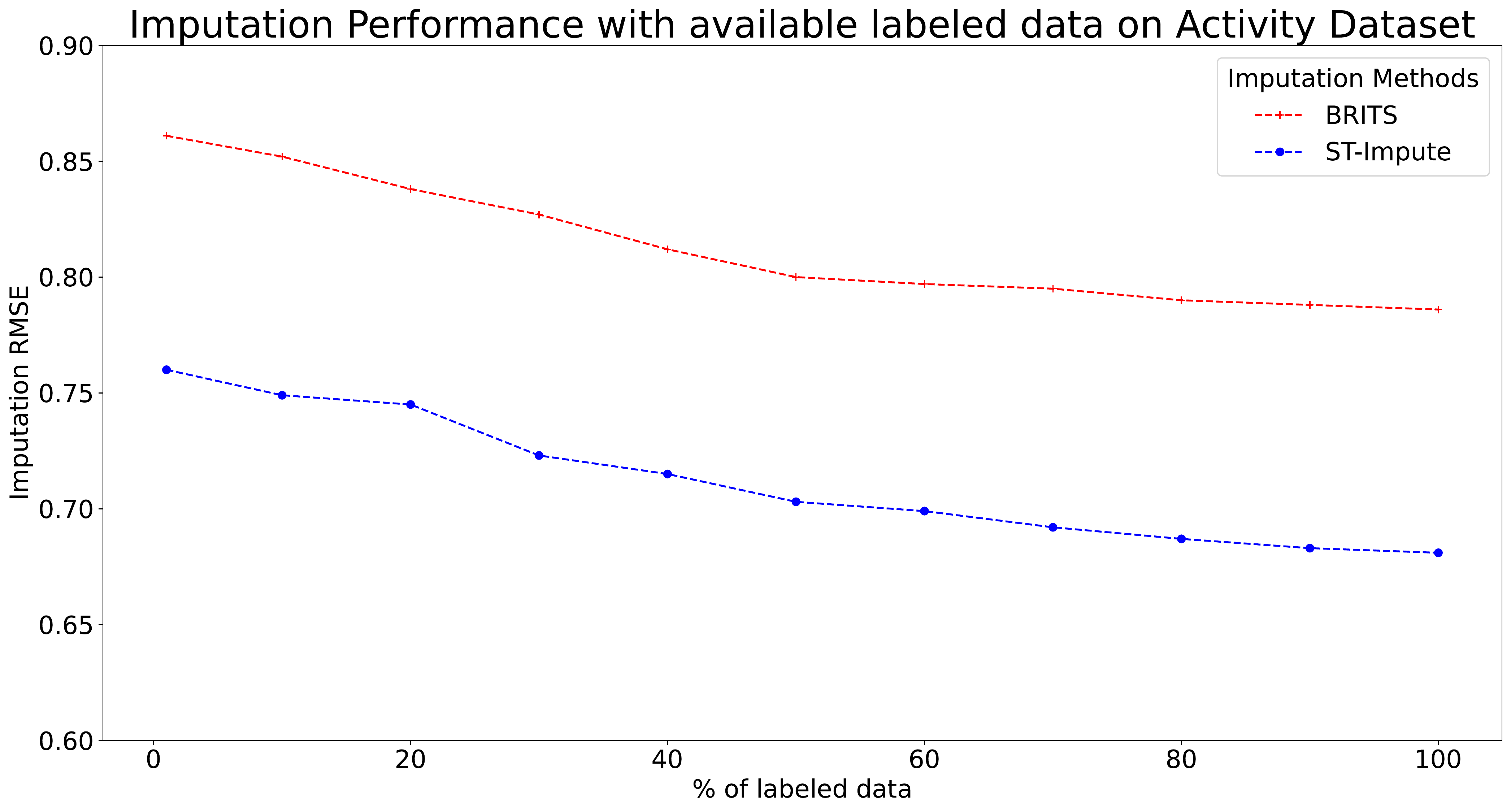}
\centering
\caption{Imputation Performance (RMSE) as we increase the number of available labels during training for the Activity Dataset.}
\centering
\label{fig:sslimpute}
\end{figure}

\subsection{Ablation study: Amount of labeled data available}
In this experiment, we perform an ablation study by observing the imputation performance as a function of the amount of available labeled data. We use the Activity dataset for this experiment as we use the classification task labels during our training. 
We compare this with BRITS, the baselines supervised learning method. Figure~\ref{fig:sslimpute} shows the results as we increase the number of available labels from 0\% to 100\%. We observe that as the number of labels increases, the imputation performance increases (RMSE decreases). This demonstrates: 

1) Imputation performance and classification labels are positively correlated; and

2) Our method consistently outperforms BRITS in a low label as well as high label data (\textbf{unlabeled}) regime.

Based on 1), we can say that the hypothesis about a better imputation helping in downstream tasks has a merit to it. This further adds to the evidence from  classification task results that ST-Impute is a better imputation method. 
Unlike methods like BRITS or GRU-D, our method is not dependent on downstream labels and hence, can give considerable boost even in unlabelled (0\%) settings.

\begin{figure}
\includegraphics[width=\linewidth]{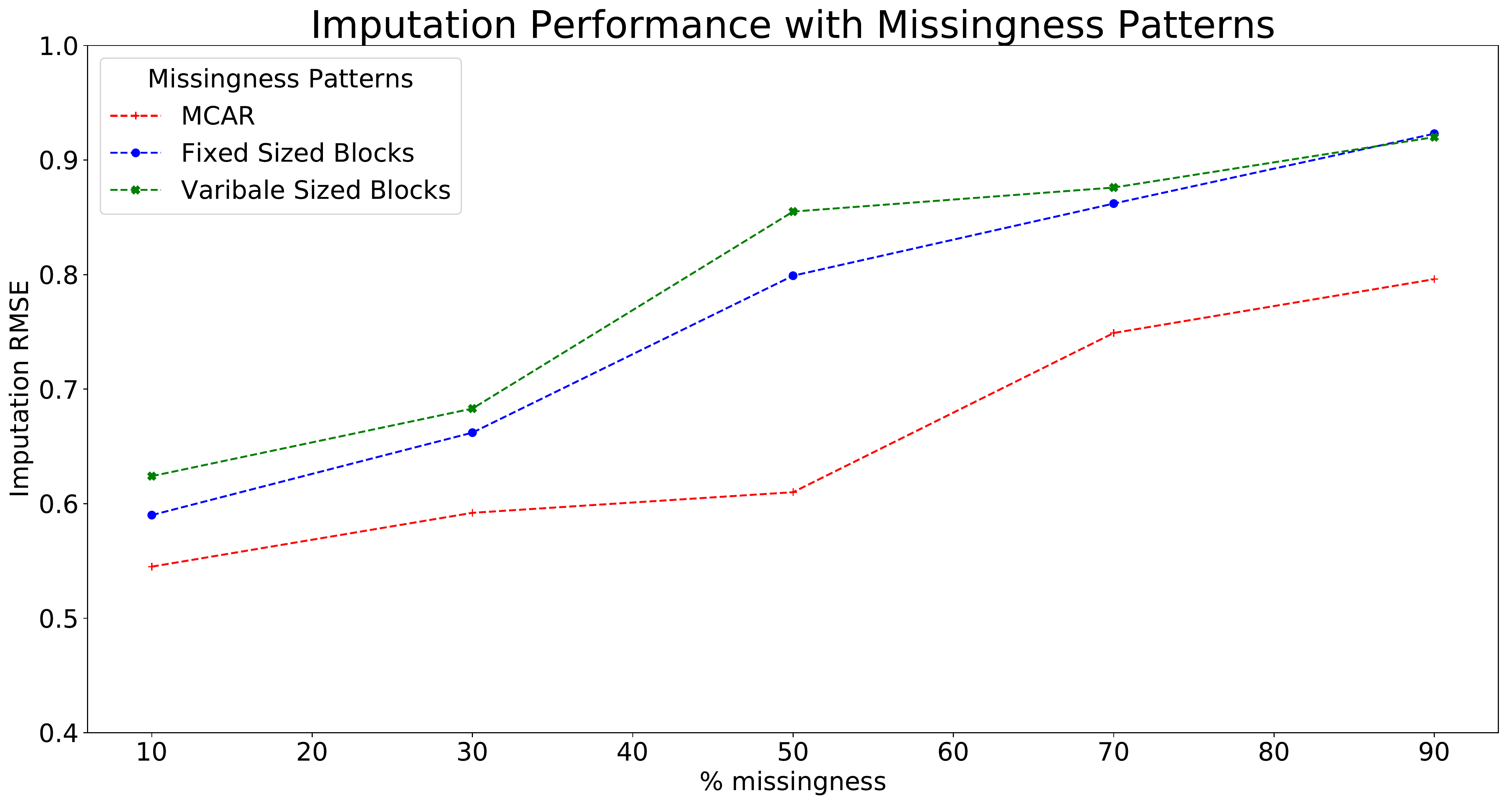}
\centering
\caption{Imputation RMSE as a function of missingness rate for Activity Dataset with ST-Impute, for three missingness patterns: MCAR, Fixed Sized missing blocks, variable sized missing blocks.}
\centering
\label{fig:missingnesspatten}
\end{figure}

\subsection{Ablation study: Missingness Pattern}
\label{sec:missingness}
Prior deep learning works have focused on using a missing completely at random (MCAR) methodology to simulate the missingness in the test set to compute the effectiveness of imputation methods. In practice, that is rarely the case. For example, for sensor data, missingness happens in continuous blocks or bursts when the sensor goes down or the data is dropped. 

In this experiment, we explore the effect of a missingness pattern on our proposed method, ST-Impute. We want to see how effective it is for imputation for completely random missing values (as reported in previous sections) versus missing blocks of values. 

We simulate a block missingness pattern with two settings, (a) fixed size blocks and (b) variable size blocks. For \emph{fixed sized blocks}, we use a block size of 10\% of time series length to create a continuous missingness pattern. For \emph{variable sized blocks}, we create blocks of length 5-15\% of time series length. Block size is sampled uniformly from [0.05,0.15] ratio interval until we reach the desired missingness percentage. 

Figure~\ref{fig:missingnesspatten} shows the imputation results with missingness rates from 10\%-90\% on the Activity dataset for ST-Impute. As we can observe, MCAR has a much lower Imputation RMSE than either of the block missingness patterns. It is expected, as with the block missingness, it becomes harder for the model to impute values as (1) any error made in imputing one value would affect immediate missing neighbouring values; (2) the missing context around a missing value becomes larger, hence degrading the performance. The widened gap at 90\% missingness rate particularly supports these observations as MCAR has at least one value in the same window to help guide imputation, while missing blocks reduce have none. 

Notice that Variable sized blocks have generally higher imputation RMSE than fixed sized blocks until 50\% missing rate. They start to converge after a 70\% missing rate, as blocks start to merge or be separated by isolated values in the middle of two missing blocks. Based on these results, it seems like the larger block sizes in the variable size blocks have more impact on the imputation quality versus the fixed sized blocks; the smaller blocks were not able to outweigh the information loss from larger blocks.

\section{Conclusions}
\label{sec3:conc}
In this work, we propose ST-Impute, a novel transformer based semi-supervised learning method to impute missing time-series values. We propose a masked imputation modeling task to train the model, which mimics time series imputation from the non-missing values. We add a diagonal self-attention mask with sparsity to enable a non-trivial time-series reconstruction on the non-missing values. Our proposed method beats existing unsupervised, supervised, and semi-supervised baselines handily on the imputation task as well as downstream tasks on three public datasets. 

\bibliographystyle{plainnat}
\bibliography{iclr2021_conference}

\end{document}